\documentclass[sigconf]{acmart}

\usepackage[toc,section=section]{glossaries}

\usepackage{booktabs} 
\usepackage{amsmath}

\usepackage{todonotes}
\usepackage{tabularx}
\usepackage{multirow}

\usepackage[linesnumbered,algoruled,boxed,lined]{algorithm2e}

\usepackage{caption}
\usepackage{subfigure}

\newcommand{\KL}[0]{\mathrm{KL}}
\newcommand{\JSD}[0]{\mathrm{JSD}}
\newcommand{\EXP}[0]{\mathbb{E}}
\DeclareMathOperator*{\argmax}{arg\,max}

\newcommand{\dataname}[1]{{\vspace{5pt}\noindent\textbf{#1}}}

\AtBeginDocument{%
  \providecommand\BibTeX{{%
    \normalfont B\kern-0.5em{\scshape i\kern-0.25em b}\kern-0.8em\TeX}}}

\copyrightyear{2019} 
\acmYear{2019} 
\setcopyright{acmcopyright}
\acmConference[SIGIR '19]{Proceedings of the 42nd International ACM SIGIR Conference on Research and Development in Information Retrieval}{July 21--25, 2019}{Paris, France}
\acmBooktitle{Proceedings of the 42nd International ACM SIGIR Conference on Research and Development in Information Retrieval (SIGIR '19), July 21--25, 2019, Paris, France}
\acmPrice{15.00}
\acmDOI{10.1145/3331184.3331232}
\acmISBN{978-1-4503-6172-9/19/07}

\begin{document}

\title{Triple-to-Text: Converting RDF Triples into High-Quality Natural Languages via Optimizing an Inverse KL Divergence}
\renewcommand{\shorttitle}{Triple-to-Text}

\author{Yaoming Zhu$^1$ \quad  Juncheng Wan$^1$ \quad Zhiming Zhou $^1$ \quad  Liheng Chen$^1$ \quad Lin Qiu$^1$ Weinan Zhang$^1$ \quad Xin Jiang$^2$ \quad Yong Yu$^1$}

\affiliation{%
	\institution{
		$^1$ Shanghai Jiao Tong University \quad
		$^2$ Noah's Ark Lab, Huawei Technologies 
	}
}
\email{{ymzhu, junchengwan, heyohai, lhchen, lqiu, yyu}@apex.sjtu.edu.cn }
\email{wnzhang@sjtu.edu.cn, Jiang.Xin@huawei.com}

\renewcommand{\shortauthors}{Trovato and Tobin, et al.}

\begin{abstract}
Knowledge base is one of the main forms to represent information in a structured way. 
A knowledge base typically consists of Resource Description Frameworks (RDF) triples which describe the entities and their relations. Generating natural language description of the knowledge base is an important task in NLP, which has been formulated as a conditional language generation task and tackled using the sequence-to-sequence framework. Current works mostly train the language models by maximum likelihood estimation, which tends to generate lousy sentences. In this paper, we argue that such a problem of maximum likelihood estimation is intrinsic, which is generally irrevocable via changing network structures. Accordingly, we propose a novel Triple-to-Text (T2T) framework, which approximately optimizes the inverse Kullback-Leibler (KL) divergence between the distributions of the real and generated sentences. Due to the nature that inverse KL imposes large penalty on fake-looking samples, the proposed method can significantly reduce the probability of generating low-quality sentences. Our experiments on three real-world datasets demonstrate that T2T can generate higher-quality sentences and outperform baseline models in several evaluation metrics. 
\end{abstract}



\maketitle

\section{Introduction}
\label{Introduction}
\begin{figure}[htbp]
	\subfigure[Knowledge base and its RDF triples.]{
		\label{KBa}
		\includegraphics[width=0.4\textwidth]{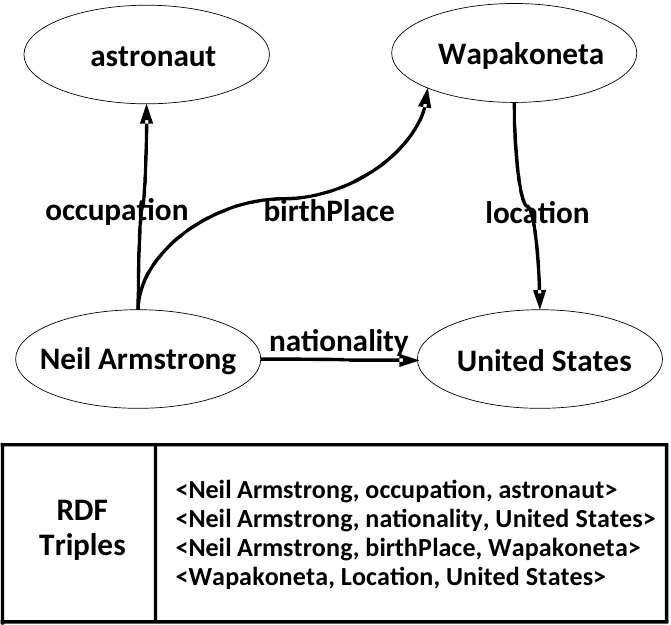}}
	\subfigure[Corresponding natural language description.]{
		\label{KBb}
		\hspace{1.5pt}
		\includegraphics[width=0.4\textwidth]{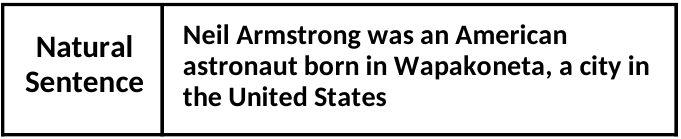}}
	\caption{A small knowledge base, (a) its associated RDF triples and (b) an example of the corresponding natural language description.}
	\label{KB}
\end{figure}
Knowledge bases (KB) are gaining attention for their wide range of industrial applications, including, question answering (Q\&A) systems \cite{fader2014open,zou2014natural}, search engines   \cite{ding2004swoogle}, recommender systems \cite{huang2002graph} etc. The \textit{Resource Description Frameworks} (RDF) is the general framework for representing entities and their relations in a structured knowledge base. Based on W3C standard  \cite{magazine1998introduction}, each RDF datum is a triple consisting of three elements, in the form of \textit{(subject, predicate, object)}. An instance can be found in Figure \ref{KBa}, which illustrates a knowledge base about Neil Armstrong and its corresponding RDF triples.

Based on the RDF triples, the Q\&A systems can answer questions such as "which country does Neil Armstrong come from?" Although such tuples in RDF allow machines to process knowledge efficiently, they are generally hard for humans to understand. Some human interaction interfaces (e.g., DBpedia\footnote{https://wiki.dbpedia.org/}) are designed to deliver knowledge bases in the form of RDF triples in a human-readable way. 

In this paper, given a knowledge base in the form of RDF triples, our goal is to generate natural language description of the knowledge bases which are grammatically correct, easy to understand, and capable of delivering the information to humans. Figure~\ref{KBb} lays out the natural language description given the knowlege base about Neil Armstrong. 

Traditionally, the Triple-to-Text task relies on rules and templates  \cite{dale2003coral,turner2010generating,cimiano2013exploiting}, which requires a large number of human efforts. Moreover, even if these systems are developed, they are often faced with problems of low scalability and inability to handle complex logic.

Recently, with significant progress on deep learning, the neural network (NN) based natural language generation models, especially the sequence to sequence framework (SEQ2SEQ)  \cite{sutskever2014sequence}, have achieved remarkable success in machine translation\cite{bahdanau2014neural} and text summarization\cite{nallapati2016abstractive}. The SEQ2SEQ framework has also been employed to translate knowledge bases into natural languages. 
Vougiouklis et al.  \cite{vougiouklis2018neural} proposed Neural Wikipedian to generate summaries of the RDF triples. 

However, most existing studies focus on the design of the model structure \cite{vougiouklis2018neural}, while paying less attention to the training objective. These models are usually trained via maximum likelihood estimation, which is equivalent to minimizing Kullback-Leibler (KL) divergence between the ground-truth conditional distribution ($P$) and the estimated distribution ($G$), i.e., $KL(P\|G)$.  Models trained with KL divergence tend to have high diversity, but at the same time,  they are likely to generate shoddy samples \cite{huszar2015not}. 

In such tasks, we usually care more about the quality of the translation and care less about diversity. Hence, we propose the triple-to-text model. By introducing a new component called \emph{judger}, we optimize the model in two directions: minimizing the approximated inverse KL divergence and maximizing the self-entropy.

Our main contributions can be summarized as follows:
\begin{list}{\textbullet}{}
	\item We propose a  theoretically sound and empirically effective framework (T2T) for optimizing the inverse KL divergence for conditional language generation task of translating a knowledge base into its natural language description.
	\item We conduct a series of experiments on different datasets to validate our proposed method. The results show that our method outperforms baselines in common metrics.
	
\end{list}

We organize the remaining parts of this paper as follows. In Section \ref{Preliminaries}, we formulate the problem and introduce the preliminaries. In Section \ref{sec_objective}, we provide our analysis of why it is preferable to optimize an inverse KL divergence. Then Section \ref{Methodology} details our proposed model. We then present the experiment results in Section \ref{Experiments}. Finally, we discuss the related work in Section \ref{Related Works} and conclude the paper in Section \ref{Conclusion}. 

\begin{table}[th]
	\caption{Glossary}
	\begin{tabular}{c|l}
		\hline
		\textbf{Symbol} & \textbf{Description} \\ \hline
		$\mathcal{F}$	&   a knowledge base that consists of RDF triples                    \\
		$t$	&        a resource description framework (RDF) triple               \\
		$\langle s_i, p_i, o_i \rangle$	&   subject, predicate and object within a RDF triple                   \\
		
		$S$	&        a sentence              \\
		$w$	&   a word in a sentence                   \\
		\hline
		$X$ 	&   conditional context for SEQ2SEQ framework     \\ 
		$Y$ 	&   target context for generative models  \\ 
		$x_i$ 	&   $i$-th token from conditional context     \\
		$y_i$ 	&   $i$-th token from target context     \\
		$y_{<i}$ 	&   prefix of target context: $\{y_1, y_2,\cdots,y_{i-1}\}$    \\
		\hline
		$P$	&   the target (ground-truth) distribution  \\
		$G_\theta$	&   learned distribution of generator \\
		$M_\phi$ 	&   learned distribution of judger     \\ $\theta$	&   parameters of generator       \\
		$\phi$	&   parameters of judger                   \\
		
		\hline
	\end{tabular}
	\vspace{4pt}
\end{table}

\section{Formulation and Preliminaries}
\label{Preliminaries}
In this section, we formulate the task and introduce the preliminaries of language generation models.

\subsection{Task Definition}
A knowledge base $\mathcal{F}$ is formulated as a set of RDF triples, i.e., $\mathcal{F} = \{ t_1, t_2, \cdots, t_N \}$, 
where each RDF triple $t_i$ is represented as $\langle s_i, p_i, o_i \rangle$. The three elements in a triple denote \emph{subject}, \emph{predicate} and \emph{object},  respectively. 
Given the knowledge base $\mathcal{F}$, our goal is to generate a natural language sentence $S$ which consists of a sequence of words $[w_1, w_2, \cdots, w_M]$, where $w_m$ denotes the $m$-th word in the sentence $S$. The generated sequence $S$ is required to be grammatically sound and correctly represent all the information contained in the knowledge base $\mathcal{F}$. 

\begin{figure*}[th]
	\centering 
	\subfigure[$P$]{
		\label{sub1}
		\includegraphics[width=0.2\textwidth]{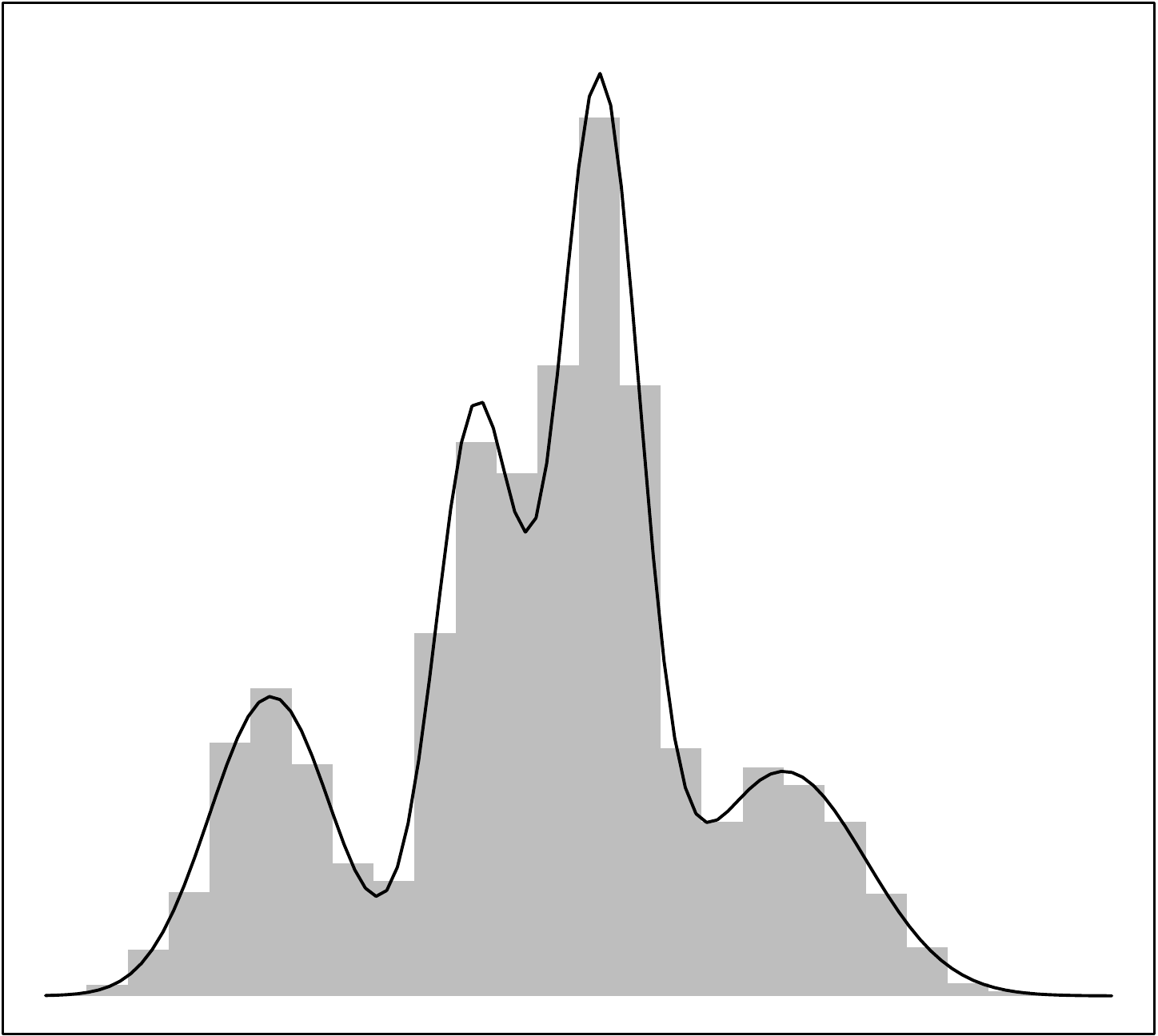}}
	\subfigure[$\KL(P\|G)$]{
		\label{sub2}
		\includegraphics[width=0.2\textwidth]{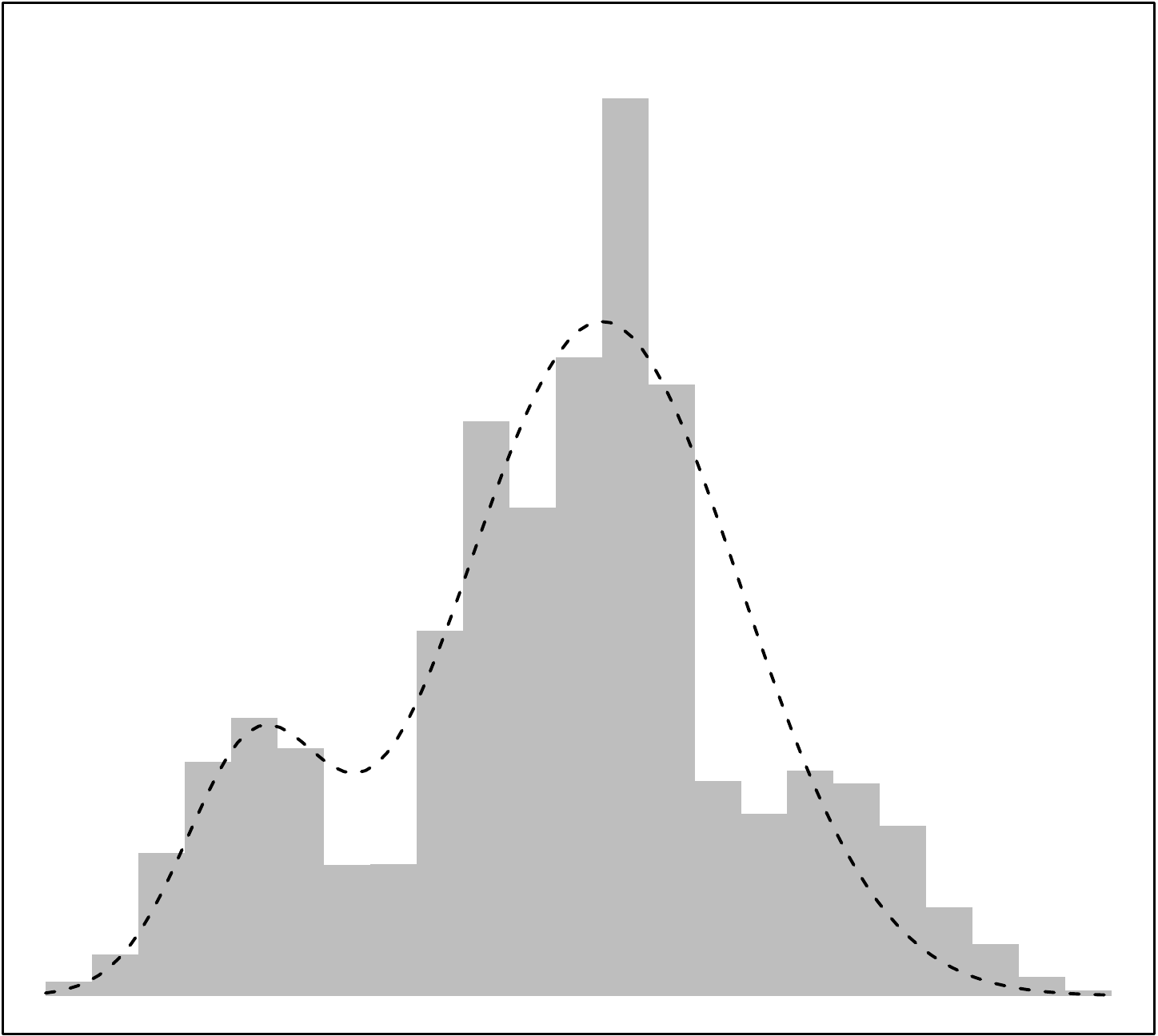}}
	\subfigure[$\KL(G\|P)$]{
		\label{sub3}
		\includegraphics[width=0.2\textwidth]{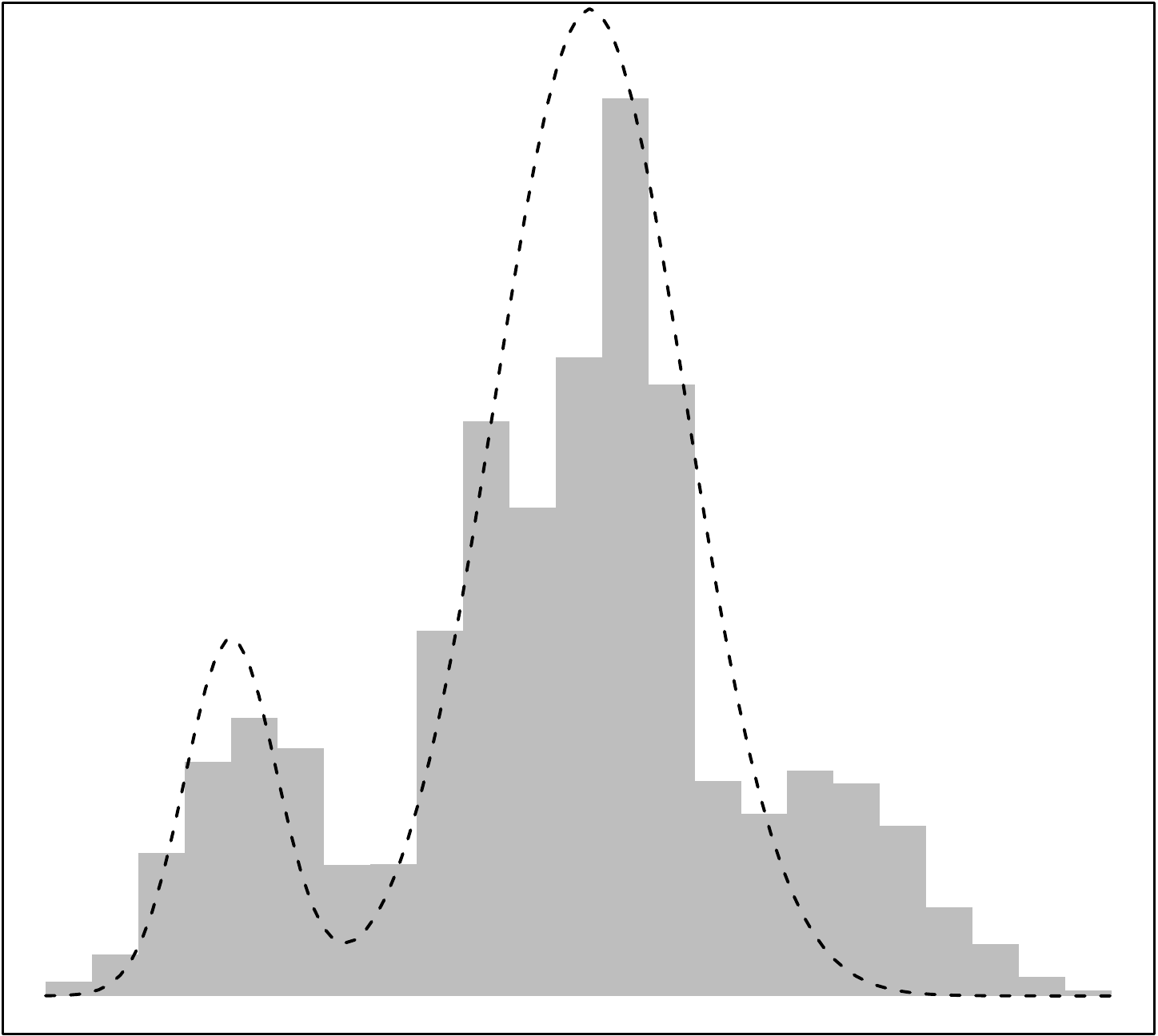}}
	\caption{ (a) shows the target distribution $P$, and the histogram in the background represents frequency of different samples; (b), (c) illustrate the empirical results of $G$ by minimizing $\KL(P\|G)$ and $\KL(G\|P)$ respectively. 
	}
	\label{main}
\end{figure*}

\subsection{Sequence to Sequence Framework}
\label{Seq2Seq}
Our work is based on the sequence to sequence framework (SEQ2SEQ). The standard sequence to sequence framework consists of an encoder and a decoder. Both of them are parameterized by \textit{recurrent neural networks} (RNN).  

The encoder takes in a sequence of discrete tokens $X = [x_1, x_2, \cdots, x_L]$. At $t$-th step, 
the encoder takes in a token and updates the hidden state recurrently: 
\begin{equation}
\mathbf{h}_t^{\text{enc}} = f^{\text{enc}}(\mathbf{h}_{t-1}^{\text{enc}}, \mathbf{e}_{x_t}),
\end{equation}
where $\mathbf{e}_{x_t}$ denotes the word embedding  \cite{mikolov2013distributed} of the $t$-th token. In general, $\mathbf{e}_{x_t} = W_e x_t$, where $W_e$ is a pre-trained or learned word embedding matrix with each column representing a embedding vector of a token; given $x_t$ is a one-hot vector, $W_e x_t$ get the corresponding column of $W_e$ for token $x_t$. $f^{\text{enc}}$ is a nonlinear function. \textit{Long short-term memory} (LSTM)  \cite{hochreiter1997long} and \textit{gated recurrent unit} (GRU)  \cite{cho2014learning} are often considered as the paradigm of the function. 
The final output of encoder is an array of hidden states $\mathbf{H}^{\text{enc}} = [\mathbf{h}^{\text{enc}}_1, \mathbf{h}^{\text{enc}}_2, \cdots, \mathbf{h}^{\text{enc}}_L]$. Each hidden state can be regarded as a vector representation of all the previous tokens. 

The decoder takes in the hidden states $\mathbf{H}^{\text{enc}}$ of the encoder as input and outputs a sequence of hidden states $\mathbf{H}^{\text{dec}}$. The hidden state at its $t$-th step is computed by: 
\begin{equation}
\mathbf{h}_t^{\text{dec}} = f^{\text{dec}}(\mathbf{h}_{t-1}^{\text{dec}}, \mathbf{e}_{y_{t-1}}, \mathbf{c}_t),
\end{equation}
where $\mathbf{e}_{y_{t-1}}$ is the word embedding of the last output token of the decoder, and $\mathbf{e}_{y_{0}}$ is set to be a zero vector. $\mathbf{c}_t$  is a function of hidden states of encoder that provides the summary of the input sequence at step $t$, and typical choices include: i) $\mathbf{c}_t=\mathbf{h}^{\text{enc}}_L$; ii) the attention mechanism  \cite{bahdanau2014neural} with $\mathbf{c}_t=g(\mathbf{h}^{\text{enc}}_1, \mathbf{h}^{\text{enc}}_2, \cdots, \mathbf{h}^{\text{enc}}_L, \mathbf{h}_{t-1}^{\text{dec}})$.

In general,  generation of language is modeled as an autoregressive sequential generation process with each token sampled from the probability distribution conditioning on its previous tokens. The probability distribution of $t$-th token is parameterized by a softmax over an affine transformation of the decoder's hidden state at step $t$, i.e.
\begin{align}
&\Pr(y_{t}| x_1, x_2, \cdots, x_L, y_{<t}) = \mathrm{softmax}(W h_t^{\text{dec}} + b),
\end{align}
where $W$ and $b$ are the weight matrix and the bias vector of output layer respectively, and  $y_{<t}$ denotes the first $t-1$ tokens of target content. 
The probability distribution of the entire output sequence $Y$ conditioning on the input sequence $X$ is thus modeled as 
\begin{align}
\Pr(Y|X) &= \Pr(y_1, y_2, \cdots, y_T | x_1, x_2, \cdots, x_L) \\\nonumber
&= \prod_{t=1}^{T} \Pr(y_t | x_1, x_2, \cdots, x_L, y_{<t}).
\end{align}

\subsection{Maximum Likelihood Estimation (MLE)}

Training neural language models through maximum likelihood estimation (MLE) is the most widely used method. 
The objective is equivalent to minimizing the cross entropy between the real data distribution $P$ and the estimated probability distribution $G_\theta$ by the generative models: 
\begin{align}
\label{KL_CE}
&J_G(\theta) =  \EXP_{Y \sim P} [ \log G_\theta(Y) ] = 
-H(P, G_\theta),
\end{align} 
where $Y$ denotes a complete sequence sampled from the real data distribution $P$ and $H$ denotes the cross entropy. 

Maximizing Eq. \ref{KL_CE} is equivalent to minimizing the Kullback-Leibler (KL) divergence between target distribution $P$ and learned distribution $G_\theta$, which is defined as 
\begin{align}
\label{MLE}
\KL(P \,\|G_\theta)  = \EXP_{Y \sim P} \big[\log\dfrac{P(Y)}{G_\theta(Y)} \big] = H(P, G_\theta) - H(P),
\end{align}
where $H(P)$ is a constant irrelevant to parameter $\theta$. 

For clarity, we here ignore the conditional context $X$ here.  
We will later regard the maximum likelihood estimation as minimizing the Kullback-Leibler (KL) divergence.

\section{Objective Analysis} \label{sec_objective}

In this section, we will give a detailed discussion on the fundamental problems of minimizing KL divergence in training and explain why we choose the inverse KL divergence as our optimization objective. We will also discuss several related solutions. 

\subsection{Practical Tendency of KL and Inverse KL}
The KL divergence between two distributions $P$ and $G_\theta$ is formulated as
\begin{equation}
\KL(P\|G_\theta)= \sum_{Y} P(Y) \log \frac{P(Y)}{G_\theta(Y)}.
\end{equation}
Since the KL divergence is non-negative, it is minimized when $G_\theta=P$. 
Unfortunately, in real-world scenarios, the target $P$ is usually a very complex distribution. Given limited capacity, the learned probabilistic model $G_\theta$ may only be a rough approximation. 

As pointed out in \cite{arjovsky2017towards}, $\KL(P\|G_\theta)$ goes to infinity if $P(Y) > 0$ and $G_\theta(Y) \rightarrow 0$, which means that the cost function is extremely high when the distribution of generator fails to cover some patterns of the real data. On the other hand, the cost function is relatively low when the generator is low-quality samples, as $\KL(P\|G_\theta)$ goes to zero if $G_\theta(Y) > 0$ and $P(Y) \rightarrow 0$.

That is, although the optimal is guaranteed to be $G_\theta=P$ under MLE objective, during training, 
the estimated distribution $G_{\theta}(y)$ is more likely to have a wide coverage and possibly contain samples out of the real distribution, as illustrated in Figure \ref{sub2}. In practice, models trained via MLE have a high probability of generating rarely-seen sequences, most of which are inconsistent with human expressions due to exposure bias \cite{bengio2015scheduled}.

With a similar argument to the behavioral tendency of $\KL(P\|G_\theta)$, it can be shown that $\KL(G_\theta\|P)$ has less penalty to "mode collapse", which means $G$ tend to generate a family of similar samples. By contrast, $\KL(G_\theta\|P)$ assigns a large penalty to fake-looking samples. The typical non-optimal estimation, as illustrated in Figure \ref{sub3}, is that it covers several major modes of the real distribution, but misses several minor modes. 

We here argue that in the conditional language generation task, especially such the triple-to-text tasks, minimizing the inverse KL divergence would be more preferred. Because, in these translation tasks, people usually care more about the quality of the generated text, rather than their diversity. In other words, it is tolerable to have low diversity, but it is usually unacceptable to be grammatically incorrect or miss important information. 
\begin{figure}[th]
	\includegraphics[width=0.90\linewidth]{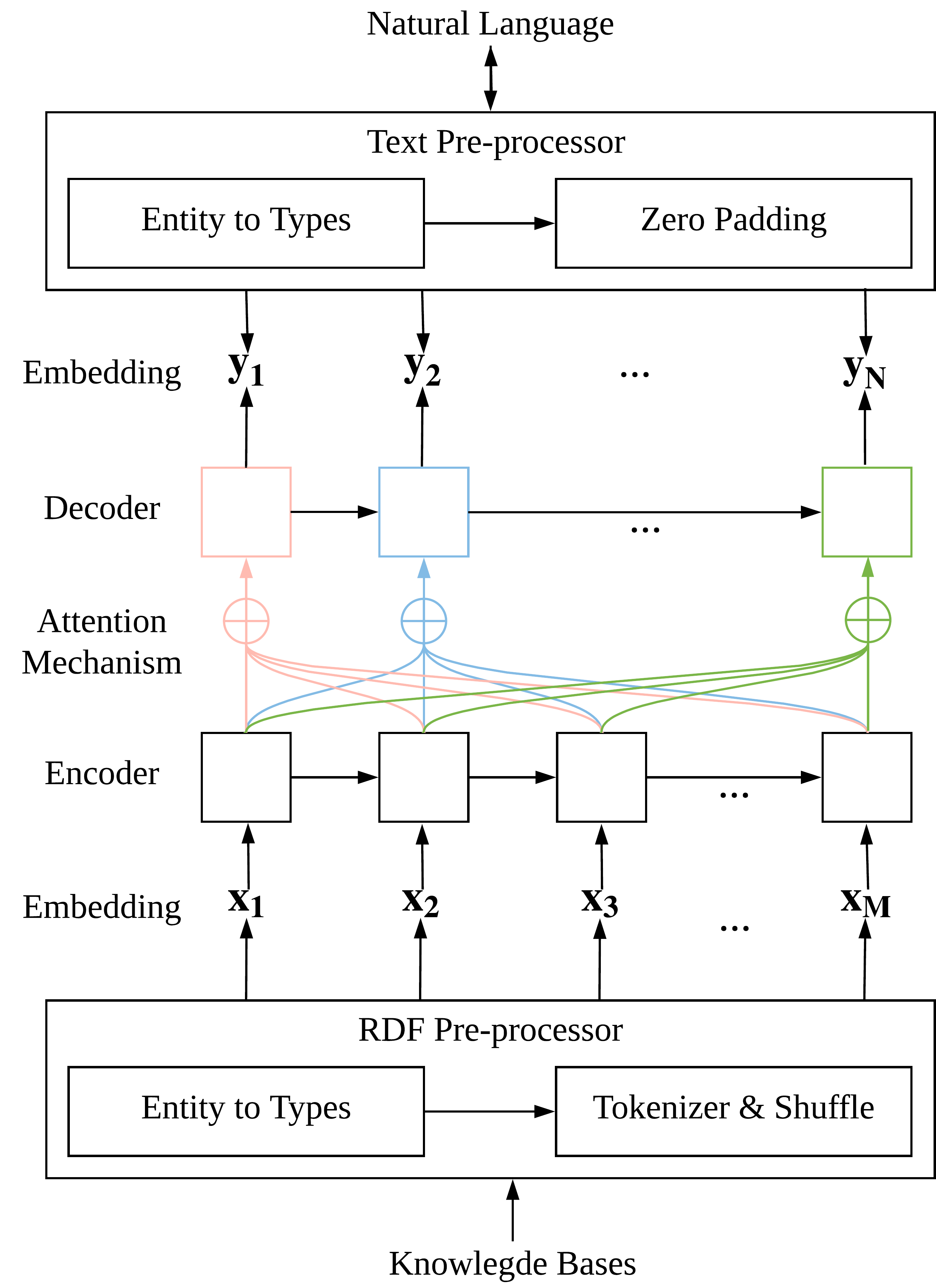} 
	\caption{General framework. Sentences and RDF triples are pre-processed into discrete tokens. Then after embedding, they are fed into an encoder-decoder neural network with attention mechanism.}
	\label{framework}
\end{figure}
\subsection{The Decomposed Objective of Inverse KL}

Here, we explain the property of inverse KL divergence via objective decomposition.  We will show that minimizing the inverse KL divergence $\KL(G_\theta\|P)$ can be regarded as a direct optimization of the performance of the Turing test.  

In Turing test , we assume that the human judges know the accurate natural language distribution $P$.\cite{mahoney1999text} Given a language sample $X$, its quality is scored by $P(X)$. Thus the averaged score in a Turing test can be modeled as the negative cross-entropy $-H(G_\theta, P) = \EXP_{Y \sim G_\theta} [ \log P(Y) ]$. 

The inverse KL divergence can be rewritten as
\begin{align}
\label{IKL}
\KL(G_\theta \,\|P ) = H(G_\theta, P) - H(G_\theta).
\end{align}
Eq. \ref{IKL} illustrates that the objective of minimizing an inverse KL divergence can be be decomposed into two parts:
\begin{list}{\textbullet}{}
	\item Minimizing $H(G_\theta, P)$, which corresponds to the objective of Turing test. 
	\item Maximizing $H(G_\theta)$, the self-entropy of the generator. It helps expand the support of $G_\theta$, to avoid disjoint support between $P$ and $G_\theta$, which may lead to gradient vanish problem  \cite{arjovsky2017towards}.
\end{list} 

\subsection{Estimation of the Real Distribution $P$}

In most real applications, $P$ is an empirical distribution and not directly accessible. For this reason we could not \emph{directly} optimize the inverse KL divergence. In our proposed method, we introduce a new module $M_\phi$, called judger, to approximate target distribution $P$. The judger is trained via maximum likelihood estimation and the objective function for $M_\phi$ is
\begin{equation}
\begin{aligned}
J_M(\phi) = \EXP_{(X, Y) \sim P}[\log M_\phi(Y|X)]. 
\label{M_update}
\end{aligned}
\end{equation}

Note that, although the judger distribution $M_\phi$ might suffer from the problems as mentioned earlier of MLE, i.e., it does not precisely model all the modes, it generally widely covers the distribution with the major modes having large probability masses. Then, based on this inaccurate estimated distribution $M_\phi$, we minimize the inverse KL divergence $\KL( G_\theta \| M_\phi)$. As we discussed before, the inverse KL divergence cares more about the major modes and tends to ignore these minor modes, including small fake modes stemming from imperfect MLE estimation, so the shortcoming of MLE-based estimated distribution $M_\phi$ poses no serious problems here.

It is also important to notice that, if the two steps in our algorithm both get the optimum, we have $G_\theta=P$, which is the same as previous methods. The key benefit of our algorithm is that when it does not get the optimum, the generated samples still tend to be feasible. 
\begin{figure*}[h]
	\centering
	\includegraphics[width=0.95\linewidth]{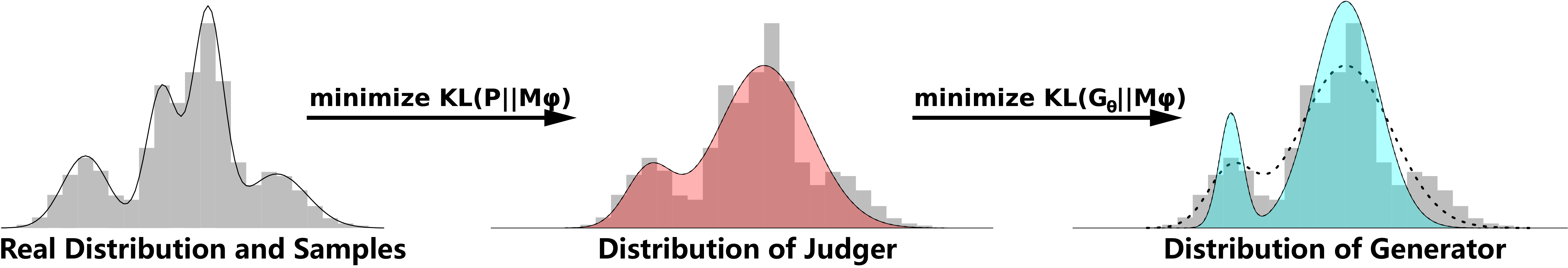}
	\caption{The overall training process of our proposed algorithm.} 
	\label{process}
\end{figure*}
\subsection{JS Divergence: GANs and CoT}
Some previous works also recognized the limitations of KL divergence and alleviated this problem with various optimization methods. Generative Adversarial Networks (GAN)  \cite{goodfellow2014generative} introduced a module named discriminator to distinguish whether a sample is from the real distribution or is forged by $G_\theta$. In theory, given a perfect discriminator, the training objective of GAN is equivalent to minimize Jensen-Shannon Divergence, which is defined as the symmetrized version of two aforementioned divergences: 
\begin{equation}
\JSD(P\|G_\theta)  = \frac{1}{2} \big(\KL(P\|M)+ \KL(G_\theta\|M)\big),
\end{equation}
where $M = \frac{1}{2} (P+G_\theta)$ is the average of two distributions. 

GAN is initially designed for generating continuous data, which is not directly applicable to discrete tokens, such as language sentences. SeqGAN  \cite{yu2017seqgan} is introduced to generate discrete tokens via adversarial training. However, the generative models trained by SeqGAN tend to have high variance due to the REINFORCE algorithm  \cite{williams1992simple}. 

Another attempt to leverage Jensen-Shannon divergence on sequence generation tasks is CoT   \cite{lu2018cot}. CoT introduces a new module, the mediator $M_\phi$, which estimates the mixture data distribution $\frac{1}{2} (P+G_\theta)$ via maximum likelihood estimation. Then $M_\phi$ is used to guide the training of the generator $ G_\theta $ with JSD. 
However, in practice, we find that the optimization of $M_\phi$ could be problematic. According to our experiments, as the real distribution becomes complicated, $M_\phi$ tends to get a distribution to fit $G_\theta$ rather than accurately modeling the $\frac{1}{2} (P+G_\theta)$. We explain this phenomenon as follows. 

The real distribution $P$ is relatively complex, and the estimated distribution $G_\theta$ tends to be simple and smooth. Because of the wide coverage tendency of MLE, $M_\phi$ would cover $\frac{1}{2} (P+G_\theta)$ in general; while due to limited capacity of $M_\phi$, $M_\phi$ tends to fit the simple one, i.e., $G_\theta$.

The problem that $M_\phi$ captures limited differences between $P$ and $G_\theta$ makes the training hard to converge. 

Note that one key difference is that the target mediator distribution in CoT is dynamical and involves with the learning distribution $G_\theta$, while the judger in our method is estimating the static distribution $P$.

\section{Methodology}
\label{Methodology}

In this section, we will first explain how to convert the task into a sequence-to-sequence generation problem, and then illustrate the details of how to optimize it with inverse KL divergence.

\SetKwRepeat{Do}{do}{while}
\begin{algorithm}
	\KwIn{a corpus of knowledge bases and its corresponding natural sentences \{($\mathcal{F}$, $S$)\}, hyper-parameters $m$ and $g$} 
	\KwOut{a generator $G_\theta$, a judger $M_\phi$ }
	Pre-process the knowledge bases corpus \{($\mathcal{F}$, $S$)\} into discrete token sequence pairs \{($X$, $Y$)\} \\
	Initialize $G_\theta$ and $M_\phi$ with random parameters $\theta$ and $\phi$ \\
	Pre-train $G_\theta$ using Maximum Likelihood Estimation (optional)\\
	\While{$G_\theta$ not converge  }
	{ \For{$m$ steps}
		{	Sample from sequence pairs \{($X$, $Y$)\} \\ 
			Update judger $M_\phi$ via maximizing $\EXP_{(X, Y) \sim P}[\mathop{\log} M_\phi(Y|X) ]$ 
		}
		\For{$g$ steps}{
			Sample conditional context $\hat{X}$ from pairs \{($X$, $Y$)\} \\
			Generate the estimated target sentence $\hat{Y}$ given $\hat{X}$ according to $G_\theta$ \\
			Update generator $G_\theta$ via minimizing $\mathop{\mathbb{E}}_{(\hat{X},\hat{Y})\sim G_\theta}\Big[\mathop{\log}\frac{G_\theta(\hat{Y}|\hat{X})}{M_\phi(\hat{Y}|\hat{X})}\Big]$ by Eq. \ref{eq_g}
		}
	}
	\Return{$G_\theta$, $M_\phi$}
	\caption{Triple-to-Text algorithm}\label{mainLoop}
\end{algorithm}

\subsection{General Framework}
\label{GeneralFramework}
The SEQ2SEQ framework cannot process graph-based data like RDF triples directly. Thus we first use a pre-processing technique similar to the one mentioned in  \cite{trisedya2018gtr}. 
It substitutes the subjects and objects in the RDF triples and their corresponding entities in the sentences into their types. 

For example, given a knowledge base [("Bill Gates", "founder", "Microsoft Corporation") , ("Microsoft Corporation", "startDate", "April 4, 1975" )] and its corresponding human-annotated natural sentence "Bill Gates founded the Microsoft Corporation in April 4, 1975". The pre-process module will map "Bill Gates," "Microsoft Corporation" and "April 4, 1975" into "PERSON," "CORPORATION" and "DATE" respectively, in both RDFs and the corresponding sentences. The pre-process can reduce the size of the vocabulary list, and improve the generalization capacity of the model so that it can handle most dates rather than just "April 4, 1975". Considering that the nodes in the knowledge bases are unordered, we also apply permutation among the triples to enhance the training data, and we believe this approach can improve the generalization capabilities of the final generation model.

The pre-processed RDF triples are then transformed into a sequence of discrete tokens. We use commas to separate elements within an RDF triple, and semicolons to separate different RDFs. For instance, the knowledge base mentioned above is turned into "PERSON, founder, CORPORATION; CORPORATION, startDate, DATE". Simultaneously, zero padding is used to fill all sequences into the same length. 

Finally, a SEQ2SEQ method introduced in Section \ref{Seq2Seq} is used to encode the processed triple and then translate it into a human-understandable sentence. To enhance the performance of the encoder-decoder model, attention mechanisms   \cite{bahdanau2014neural} are used in our proposed framework. Figure \ref{framework} illustrates the general structure of this method.
\begin{table*}[th]
	\caption{Dataset statistics, including the number of RDF triples-sentence pairs used in training and test, the number of RDF triples per datum, the (maximum) number of tokens per sentence and the vocabulary list size. }
	\begin{tabular}{c|cccccc}
		\hline
		Dataset   & \#Train & \#Test & \#RDF Triples & \#sentence length & \#vocabulary in sentence & \#vocabulary in triples \\ 
		\hline
		WebNLG    & 20288   & 2240   & 1-7       &        82           &      4678                    &             2718            \\ 
		SemEval   & 8000    & 2717   & 1         &      97             &       24986                   &         7333                \\ 
		Baidu SKE &    19520     &    2000    & 1-5       &      84             &         25027                 &           22713              \\ \hline
	\end{tabular}
	\label{data}
\end{table*}
\subsection{Algorithm Details} \label{sec_proposed_method}
The general idea of the proposed method is that:
a module $M_\phi$ called judger is introduced to approximate the target distribution $P$, which is trained via maximum likelihood estimation. 
Based on the approximated distribution $M_\phi$, we then minimize the inverse KL divergence $\KL(G_\theta\|M_\phi)$. The overall process is illustrated in Figure \ref{process}.

Because we target at the sequence to sequence translation task, the distribution of generator is modeled as a chain product of probability distribution of the next token $y_t$ conditioning on the input sequence $X$ and prefix $y_{<t}$,
\begin{equation}
\begin{aligned}
G_\theta(Y|X) &= \prod_{t=1}^{T} g_\theta(y_t|y_{<t}, X) .
\label{the_eq2}
\end{aligned}
\end{equation}
Within our framework, the judger is trained to model the target distribution via maximum likelihood estimation. The judger $M_\phi$ is also modeled as a chain product of conditional distributions,  
\begin{equation}
\begin{aligned}
M_\phi(Y|X) &= \prod_{t=1}^{T} m_\phi(y_t|y_{<t}, X) .
\label{the_eq22}
\end{aligned}
\end{equation}
The objective function for $ M_\phi $ is
\begin{equation}
\begin{aligned}
J_M(\phi) = \EXP_{(X, Y) \sim P}[\log M_\phi(Y|X)] .
\label{M_update}
\end{aligned}
\end{equation}
Given $M_\phi$ which estimates the real distribution $P$, we then update $G_\theta$ via minimizing the inverse KL divergence $\KL( G_\theta \,\| M_\phi)$:
\begin{align}
\label{KL_2}
J_G(\theta) &= \KL( G_\theta \,\| M_\phi) =  \mathop{\mathbb{E}}_{(X,Y)\sim G_\theta}\Big[\mathop{\log}\frac{G_\theta(Y|X)}{M_\phi(Y|X)}\Big], \end{align}
where $ (X,Y)\sim G_\theta $ denotes the data pair where $ X $ is sampled from conditional context and $ Y $ is the output of generator given $ X $ as input. The objective can be directly optimized by taking Eq.~(\ref{the_eq2}) and ~(\ref{the_eq22}) into Eq (\ref{KL_2}), which can be reformulated as 
\begin{equation}
\label{eq_g}
\mathop{\mathbb{E}}_{(X,Y) \sim G_\theta} \Big[\mathop{\sum}_{t=1}^{T} \big(\mathop{\log} g_\theta(y_t|X, y_{<t})-\mathop{\log} m_\phi(y_t|X, y_{<t}) \big) \Big]. 
\end{equation}

Algorithm \ref{mainLoop} illustrated the overall algorithm of our proposed method. 
Note that instead of training the judger to convergence at the beginning, the judger and the generator are trained alternately. From the perspective of curriculum learning \cite{bengio2009curriculum}, by gradually increasing the complexity of the generator's training objective, it improves the generalization ability of the generator and helps find a better local optimum. Our method shares the same computational complexity as MLE training.

\section{Experiments}
\label{Experiments}

\subsection{Datasets}
Our methods are evaluated on the following datasets. 

\dataname{WebNLG}   \cite{gardent2017webnlg} is extracted from 15 different DBPedia   \cite{auer2007dbpedia} categories, which consists of 25,298 (data, text) pairs and 9,674 distinct data units. The data units are sets of RDF triples, and the texts are sequences of one or more sentences verbalizing these data units. It also provides a set of 373 distinct RDF properties.

\dataname{SemEval-2010 Task 8}  \cite{hendrickx2009semeval} was originally designed for multi-way classification of semantic relations between pairs of nominals. It contains 10,717 samples, divided as 8,000 for training and 2,717 for testing. The dataset contains nine relation types. Since each example is a sentence annotated for a pair of entities and the corresponding relation class for this entity pair in this dataset, we can extract an RDF triple from each sentence.

\dataname{Baidu SKE}\footnote{http://ai.baidu.com/broad/introduction} is a large-scale human annotated dataset with more than 410,000 triples in over 200,000 real-world Chinese sentences, bounded by a pre-specified schema with 50 types of predicates. Each sample in SKE contains one sentence and a set of associated tuples. SKE Tuples are expressed in forms of (subject, predicate, object, subject type, object type). In our experiments, we only use knowledge bases related to \textit{Film and TV works} domain, and each Chinese character is treated as a distinct token.

We select some data in the three data sets and divide them into a training set and a test set.
Table \ref{data} shows some statistical details about the data.

\subsection{Implementation Details}

The generator consists of a word embedding matrix, an encoder, a decoder, and the output layer. For the word embedding, we maintain two different sets of embeddings for encoder and decoder respectively; both are of 64 dimensions. 
Both encoder and decoder are built as an LSTM   \cite{hochreiter1997long} with hidden units of 128 dimensions. The dimension of hidden units of the output layer is also 128. 
We apply Bahdanau attention \cite{bahdanau2014neural} to the context vector $c_t$, which is computed as the weighted sum of encoder states. 
For the judger, we use the same configuration as the generator.

For the initialization, all initial parameters follow a standard Gaussian distribution $\mathcal{N}(0,1)$. 
All models are optimized using Adam optimization   \cite{kingma2014adam} with a learning rate of 0.001 and a batch size of 64. 
The hyper-parameters of $g$ and $m$ in the algorithm are both set as 1, which makes the objective of the generator gradually harder as indicated in Section \ref{sec_proposed_method}. 
We also pre-train the generator via MLE with the number pre-train epochs set as 2. 

\begin{table*}[]
	\caption{Comparison of model performance. }
	\begin{tabular}{c|ccc|ccc|ccc|ccc}
		\hline
		\multirow{2}{*}{\textbf{}} & \multicolumn{3}{c|}{BLEU-3 \textuparrow} & \multicolumn{3}{c|}{BLEU-4 \textuparrow} & \multicolumn{3}{c|}{TER \textdownarrow } & \multicolumn{3}{c}{METEOR \textuparrow} \\ \cline{2-13} 
		& WebNLG   & SemEval  & SKE   & WebNLG   & SemEval  & SKE   & WebNLG & SemEval & SKE   & WebNLG  & SemEval  & SKE    \\ \hline
		MLE                        & 40.8     & 4.24     & 18.6  & 30.2     & 2.73     & 15.6  & 0.497  & 1.07   & 1.01  & 0.636   & 0.222    & 0.349  \\
		CoT                        & 9.84     & 1.90     & 15.7  & 6.40     & 1.41     & 12.9  & 1.085  & 1.16   & 1.08  & 0.349   & 0.102    & 0.305  \\
		SeqGAN                     & 42.0     & 4.11     & 19.0  & 24.4     & 2.63     & 14.1  & 0.534  & 1.11   & 1.10  & 0.597   & 0.231    & 0.344  \\
		PG                         & 41.7     & 4.21     & 17.9  & 30.9     & 2.06     & 14.1  & 0.607  & 1.13   & 1.12  & 0.628   & 0.197    & 0.310  \\
		NW                         & 35.8     & 2.80     & 14.6  & 24.6     & 1.87     & 11.9  & 1.664  & 1.17   & 1.92  & 0.302   & 0.143    & 0.301  \\\hline
		T2T               & \textbf{42.4}     & \textbf{4.35}     & \textbf{20.3}  & \textbf{32.2}     & \textbf{2.83}     & \textbf{17.1}  & \textbf{0.473}  & \textbf{0.957}   & \textbf{0.947} & \textbf{0.641}   & \textbf{0.247}    & \textbf{0.367}  \\ \hline
	\end{tabular}
	\label{bleu}
\end{table*}

\subsection{Baseline Algorithms}
We validate our proposed method for RDF triple-to-text (we will later refer to as T2T) by comparing it with the following baselines. To give a fair comparison, we apply the same RDF pre-processing technique discussed in Section \ref{GeneralFramework} to all the baselines.

\begin{list}{\textbullet}{}
	\item \textbf{MLE}. A common method for training sequence to sequence framework. For a fair comparison, the parameter setting of the generator is the same with our model.
	\item \textbf{CoT}. We adapt CoT  \cite{lu2018cot} into conditional sequence generation task. As its authors suggested, the size of the hidden unit of the mediator is twice the size of the generator. 
	\item \textbf{Pointer-Generator Network (PG)}. See et al.   \cite{see2017get} proposed pointer-generator network. Their work can be regarded as a combination of SEQ2SEQ and pointer network   \cite{vinyals2015pointer}.
	\item \textbf{SeqGAN}. Yu et al.   \cite{yu2017seqgan} used an adversarial network to provide the reward and train a sequence generator with policy gradient. According to   \cite{li2017adversarial} and our initial experiments,  in SEQ2SEQ framework, when the discriminator is parameterized as a convolutional neural network, it is difficult for the discriminator in SeqGAN to improve the generator. We thus follow   \cite{li2017adversarial} and adapt the discriminator into a hierarchical recurrent neural network   \cite{li2015hierarchical}.
	\item \textbf{Neural Wikipedian (NW)}. Vougiouklis et al.  \cite{vougiouklis2018neural} used a standard feed-forward neural network to encode RDF triples. Then the vectors derived from encoders are concatenated and used as the input of the decoder which generates summaries for RDF triples.
\end{list}

\begin{table}[h]
	\caption{predicate accuracy on SemEval dataset.}
	\begin{tabular}{c|ccccc|c}\hline
		methods  & MLE     & CoT   & PG & SeqGAN & NW & T2T  \\ \hline
		accuracy & 0.240  & 0.258 &    0.231  & 0.244 & 0.155 & \textbf{0.276} \\    \hline
	\end{tabular}
	\label{acc}
\end{table}

\subsection{Metrics}

For natural language generation tasks, the most widely accepted metric is human evaluation   \cite{belz2006comparing}.
While human evaluation is reliable, it is hardly applied to quality evaluating of large corpus since it will involve too many human resources. Therefore, we have to introduce automatic metrics for evaluating all the sentences our system has generated. However, to our knowledge, no single automatic evaluation metric is sufficient to measure the performance of a natural language generation system   \cite{novikova2017we}. Thus, in order to give objective results, we use a variety of automatic metrics to compare our models and benchmarks.

We have adopted three widely used word-level metrics:
BLEU \cite{papineni2002bleu}, TER \cite{snover2006study} and METEOR \cite{banerjee2005meteor}. BLEU and METEOR\footnote{We use METEOR 1.5 (https://www.cs.cmu.edu/~alavie/METEOR/README.html), with parameters suggested by Denkowski et al.   \cite{denkowski2014meteor} for universal evaluation} to calculate the number of $n$-grams of the generated sentence occurs within the set of references. 

Besides the traditional word-based metrics, we also evaluate the generator via likelihood and perplexity. Inspired by likelihood-based discrimination\cite{mclachlan2004discriminant}, we design a new metric which we refer to as "predicate accuracy". In detail, given a single RDF triple $X_p = \langle s, p, o \rangle$, a natural sentence $Y$ describing the triple and a generative model $G_\theta$, we can calculate $G_{\theta}(Y| X_p)$, i.e. the predictive likelihood of the target sentence. 
If we keep subject $s$ and object $o$ unchanged, and substitute predicate $p$ with another predicate $p'$, then our generative model can derive a probability density $G_{\theta}(Y| X_{p'} ) $ for each predicate class $p'$, where $X_{p'}$ denotes the triple $\langle s, p', o \rangle$. Then, we can use the likelihood of generative model to predict the predicate given subject $s$,  object $o$ and sentence $X_p$, the predicted predicate $\hat{p}$ is 
\begin{equation}
\label{accPre}
\hat{p} = \argmax_{p_i \in P} G_{\theta}(Y| X_{p_i})
\end{equation}
where $P$ is the set of all kinds of predicate. The "predicate accuracy" is defined as precision of $\hat{p}$ in Eq. \ref{accPre} being the correct predicate describing the sentence $Y$.

We also use \emph{forward perplexity} ($\mathtt{FPPL}$) to evaluate the quality of the generated text. Different from the traditional perplexity evaluated only on generative models, $\mathtt{FPPL}$ evaluate perplexity of generated samples from generator $G_\theta$ using another language model (denoted as $H_\psi$ ) trained on real data via MLE. According to Zhao et al. \cite{kim2017adversarially}, $\mathtt{FPPL}$ measures the fluency of generated sentences. 
%
\begin{equation}
\mathtt{FPPL(G_\theta)} = e^{- \EXP_{y \sim G_\theta} \log{H_\psi(y)}} 
\end{equation}
In our experiments, $H_\psi$ is implemented as an LSTM-based SEQ2SEQ model, whose word embedding size is set as 64, encoder hidden unit and decoder hidden unit is all set as 300.

\begin{table}[th]
	\vspace{5pt}
	\caption{Forward perplexity among three datasets.}
	\begin{tabular}{c|ccc} \hline
		methods & WebNLG & SemEval & SKE   \\\hline
		MLE     & 1.810  & 2.918   & 7.046 \\
		CoT     & 2.423  & 2.579   & 6.643 \\
		SeqGAN  & 3.556  & 4.129   & 7.834 \\
		PG      & 2.151  & 3.180   & 8.078   \\
		NW      & 2.461  & 2.771   & 6.916   \\
		\hline
		T2T     & \textbf{1.589}  & \textbf{2.067}   & \textbf{3.565} \\ \hline
	\end{tabular}
	\label{FPPL_table}
\end{table}

\begin{figure*}[th]
	\centering 
	\subfigure[WebNLG]{
		\label{FPPL_1}
		\includegraphics[width=0.32\textwidth]{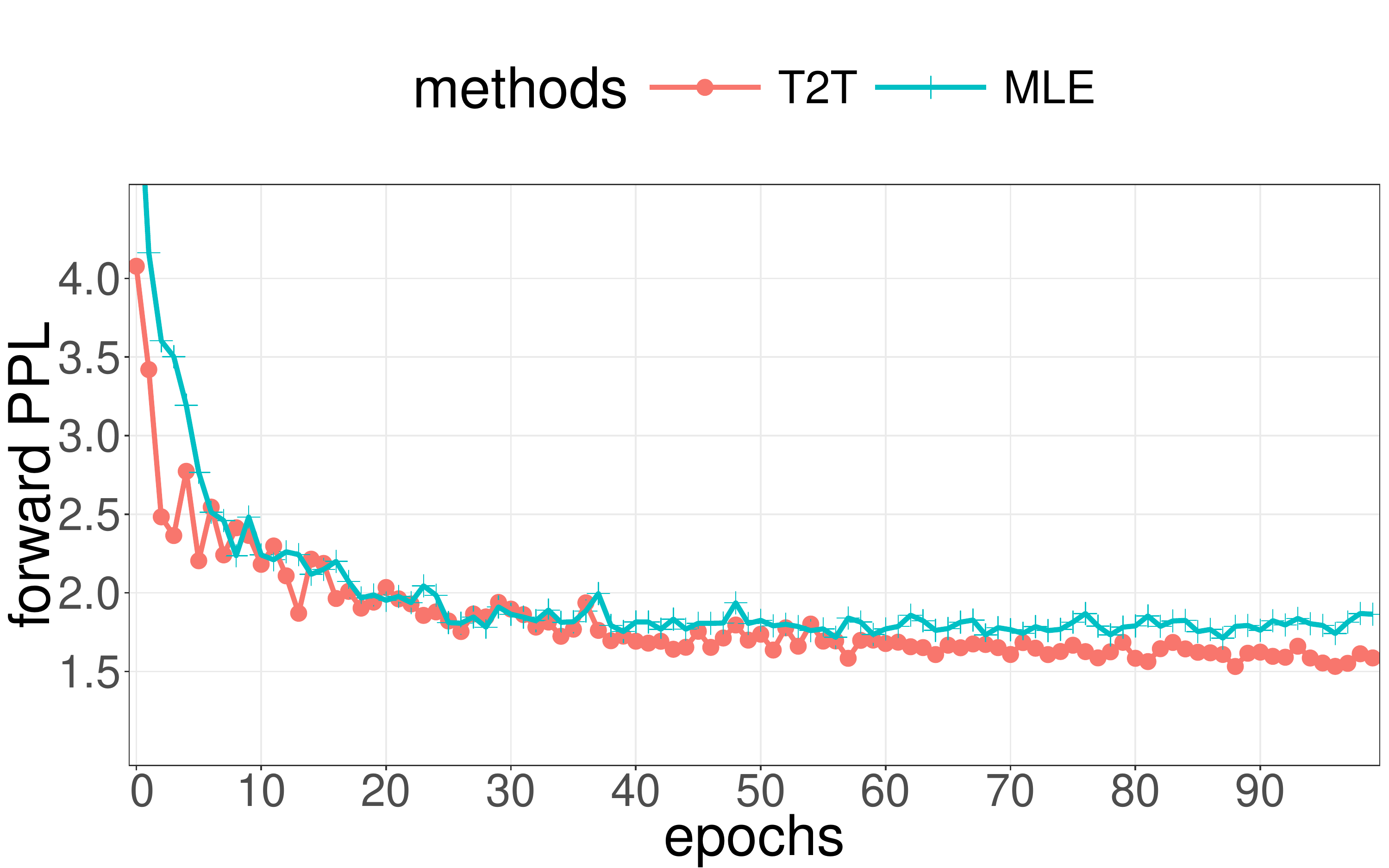}}
	\subfigure[SemEval]{
		\label{FPPL_T}
		\includegraphics[width=0.32\textwidth]{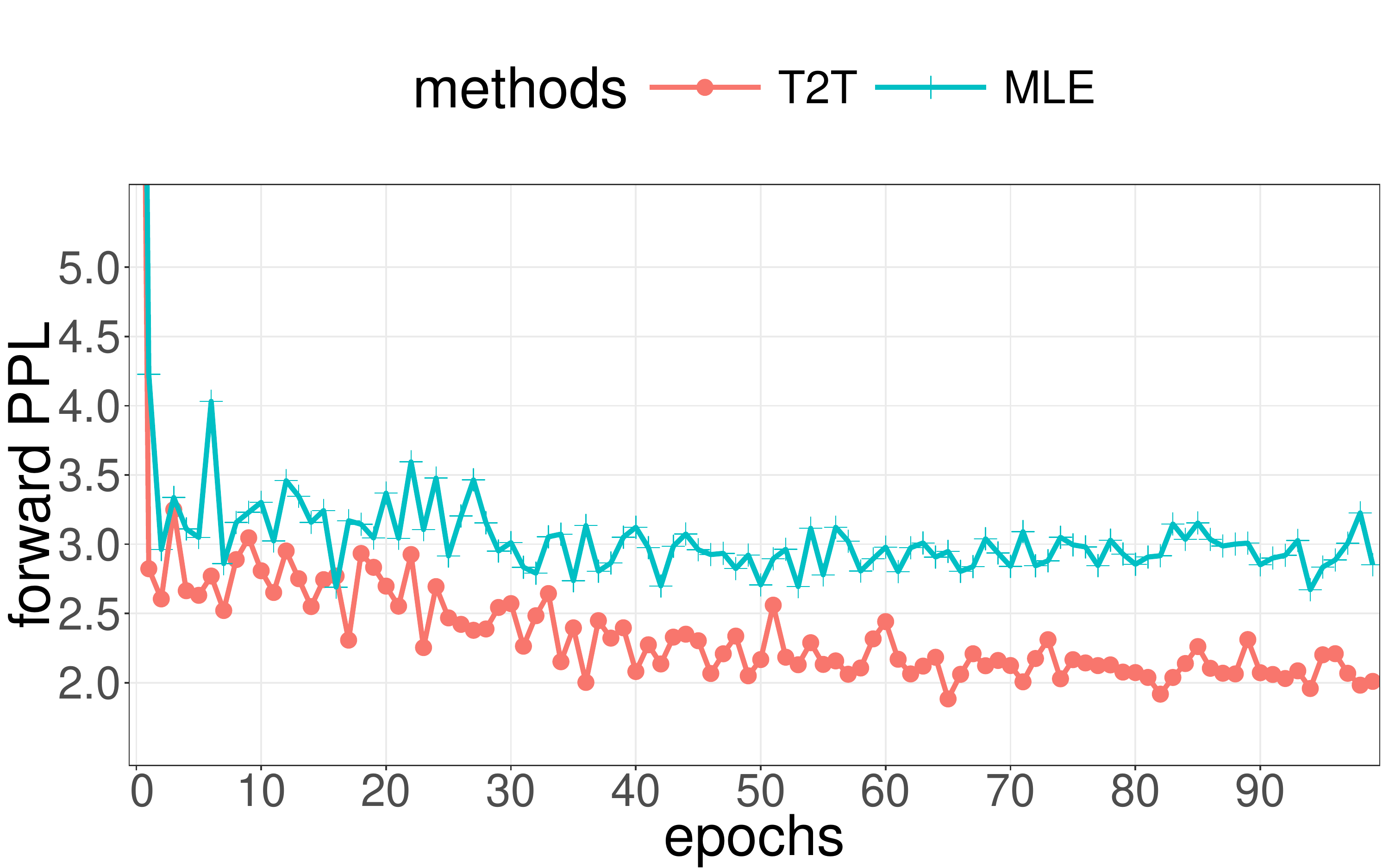}}
	\subfigure[BaiduSKE]{
		\label{FPPL_3}
		\includegraphics[width=0.32\textwidth]{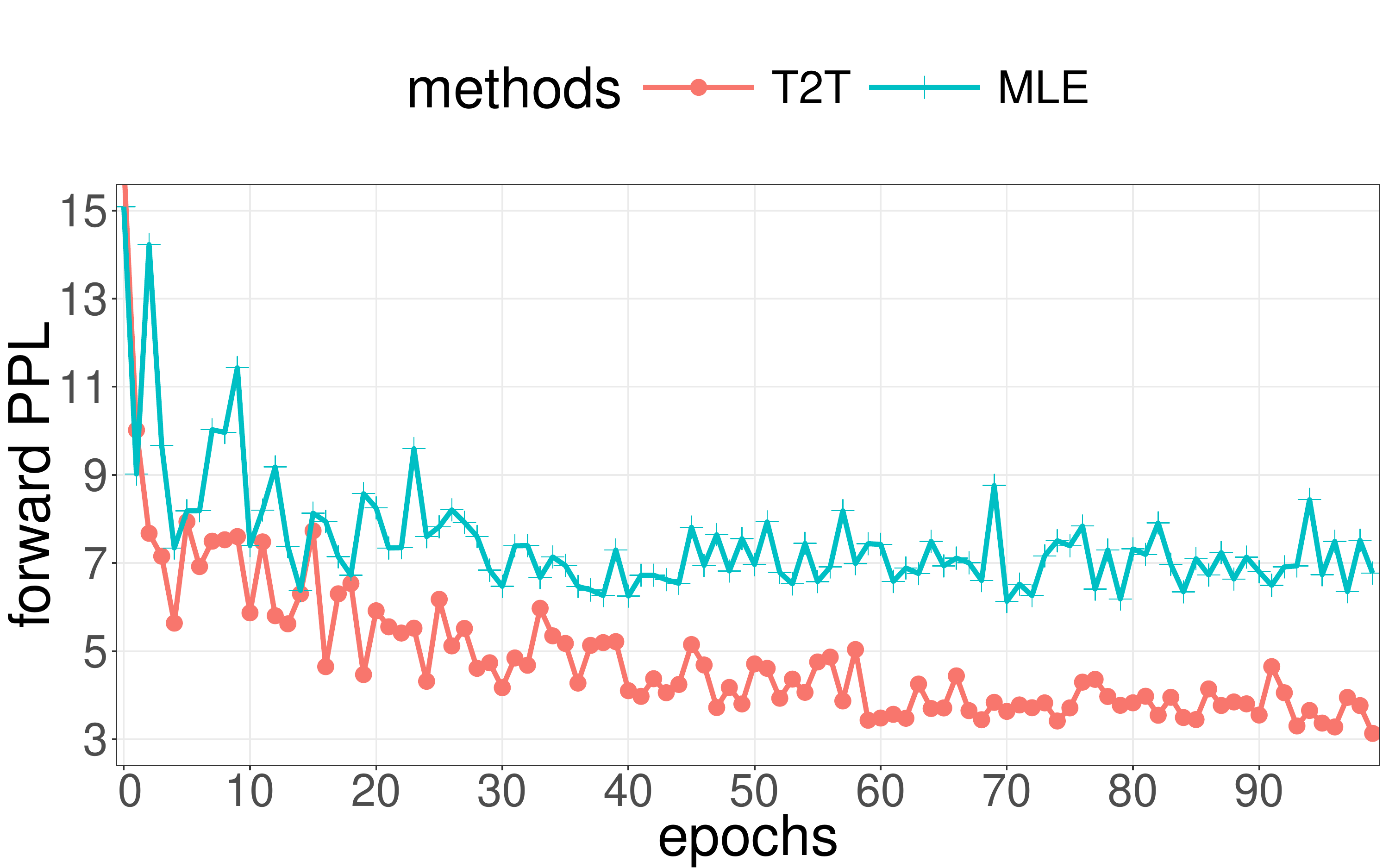}}
	\caption{ Forward perplexity training curves on Three datasets respectively.
	}
	\label{main}
\end{figure*}

\subsection{Experiments Results}
Table \ref{bleu} shows the overall results of each training method on BLEU, TER, and METEOR among the three datasets. From the results, we find that our proposed T2T method improves the quality of the generated sentence on these word-based metrics. 
As we have analyzed, generators optimized via inverse KL divergence tend to generate text with more common expressions, while other baselines tend to use some low-quality text. 
Thus, sentences from T2T will overlap more words with reference text, which means it can achieve better performance on the word based metrics like BLEU. 

The experiments on $\mathtt{FPPL}$ also validated our conclusion.  Table \ref{FPPL_T} shows the results of $\mathtt{FPPL}$ on WebNLG dataset. Low forward perplexity validates that our method allows the generator to generate high-frequency language patterns more and better. We plot the training curves of $\mathtt{FPPL}$. 

Table \ref{acc} shows the predicate accuracy of different training methods on SemEval datasets. Our model can fit the logical connection between real sentences and RDF predicates better compared with baselines. 




Human evaluation is conducted on WebNLG dataset to validate the performance of our framework further. We choose WebNLG dataset because it consists of more RDF triples and its reference sentences are relatively simple. 
We randomly select 20 RDF triples from the dataset, along with the corresponding sentences generated by T2T and baselines.
Ten human volunteers are asked to rate the sentences from two aspects: grammar and correctness. The score on grammar is used to judge whether the sentence contains grammatical errors, improper use of words and repetition. Correctness measures whether the sentence accurately represents the information in the RDF triples. The score for each criterion takes an integer between 1 and 10. Volunteers are given both scoring criteria and examples. Table \ref{human} lists the results of overall human evaluation score.

Table \ref{Samples} presents samples of generated sentences from different baselines and T2T given a knowledge base about Amatriciana sauce. 
Compared with baselines, the text generated from T2T is not only grammatically sound and correctly expresses all the information from RDF triples as well. We also found that when the length of the generated sentence is long, the quality of output from SeqGAN is compromised, which may because they use Monte Carlo sampling to guide the generator, which will introduce variance. The sentences generated by MLE correctly express the knowledge, but the grammar and the words are not quite authentic. Text generated using Pointer-Generator suffers from repetition. Neural Wikipedian can hardly express all information soundly given multiple triples.
\begin{table}[th]
	\caption{Human evaluation on WebNLG dataset.}
	\begin{tabular}{c|cc}
		\hline
		methods                  & Grammar & Correctness \\ \hline
		MLE               & 7.6     & 6.7         \\
		CoT               & 5.5     & 3.5         \\
		PG & 7.1     & 5.6         \\
		SeqGAN            & 8.0     & 5.8        \\
		NW            & 6.3     & 4.6      \\\hline
		T2T               & \textbf{8.6}     & \textbf{7.1} \\ \hline
	\end{tabular}
	\label{human}
\end{table}

\begin{table*}[]
	\caption{Sample output of the system.}
	\begin{tabularx}{0.95\textwidth}{c|X}
		\hline 
		\multirow{2}{*}{\textbf{RDF inputs}}        & <Italy , capital , Rome>, <Italy , leaderName , Matteo Renzi>, <Amatriciana sauce , country , Italy>,\ \ \ \  \ \ \ \ \ \ \ \ \ \ \ \ \ \ \ \ \ \ \ \ \ \ \ \ \ \ \ \ \ \ \ \ \ \ \ \ \ \ \ \ \ \ \ \ \ \ \ \ \ \ \ \ <Italy , leaderName , Laura Boldrini>           \\ \hline
		\multirow{2}{*}{\textbf{Reference}}      & Amatriciana sauce is a traditional sauce in italy ( the capital of which is rome ) , where two of the country ' s leaders are matteo renzi and laura boldrini . \\ \hline
		\multirow{2}{*}{MLE}               & Italy is called a country Amatriciana sauce . Matteo Renzi and Laura Boldrini are leaders in Italy where the capital is Rome .                                 \vspace{2pt} \\ 
		\multirow{1}{*}{CoT}           & Laura Boldrini is a leader in Italy where Rome is the capital of the country of Italy where where valencia is bacon.                                            \vspace{2pt} \\ 
		\multirow{2}{*}{SeqGAN}            &        Amatricana sauce comes from Italy , a political leader  and the capital is Rome . matteo renzi and Laura Boldrini are one of the leaders of Italy is                                                                                                                                                  \vspace{2pt}      \\            
		\multirow{2}{*}{PG} & Amatriciana sauce , a traditional italian dish from the Rome of the italian , where Rome the the leader is either two leaders include Matteo Renzi.               \vspace{2pt}      \\
		\multirow{1}{*}{NW} & the leader of Italy is Laura Boldrini where amatriciana sauce can be found .                \vspace{2pt}      \\\hline
		\multirow{2}{*}{T2T}           & Amatriciana sauce is from the country of Italy where capital is Rome . its leader is Laura Boldrini and Matteo Renzi leads the country .                        \\ \hline
	\end{tabularx}
	\vspace{10pt}
	\label{Samples}
\end{table*}

\section{Related Works} 
\label{Related Works}

Our task can be regarded as a combination of two problems. One is on the training of neural language models; another is on converting knowledge bases (structured data) into natural languages.

\subsection{Knowledge Base to Natural Language}
Previous approaches on generating natural language from knowledge bases can be categorized into the following types: rule-based, template-based and neural language model based.

Generating sentences based on knowledge bases with hand-crafted rules is the main technology in traditional NLG systems, which often involves domain-specific knowledge and only works for a particular data type. Bontcheva et al.   \cite{bontcheva2004automatic} designed a set of rules to generate natural language reports from medical data automatically.  O'Donnell et al.   \cite{o2000optimising} designed a text generation system by utilizing the potential rules from relational databases. They specified the semantics of relational databases and reconstructed an \textit{"Intelligent Labelling Explorer"} (ILEX) system. Based on that, the ILEX system can interpret entities from databases based on information like domain taxonomy and specification of the data type. Cimiano et al.  \cite{cimiano2013exploiting} presented a principled language generation architecture by analyzing statistical information derived from a domain corpus. Their system can write recipes based on RDF representations of a cooking domain. They mainly focus on extracting lexicon and then formulate the recipes with a parse tree.

Template-based generation is another traditional approach to convert structured data into text. In general, developing such kind of system often requires complex design about grammar, semantic and lexicalization \cite{deemter2005real}.
Kukich   \cite{kukich1983design} designed a knowledge-based report generator which infers semantic messages from the data and then maps that information into a grammar-based template. 
Flanigan et al.   \cite{flanigan2016generation} proposed a two-stage method for natural
language generation from Abstract Meaning Representation   \cite{banarescu2013abstract}.
Duma et al.   \cite{duma2013generating} formulated a system which automatically
learns sentence templates using the corpus extracted from Simple English Wikipedia and DBpedia. 

The former two technologies have good availability, reliability and do not rely on large quantities of corpora to train the model. However, they require a labor expert and have poor scalability.

\subsection{Neural Language Models}
Sequence-to-sequence model   \cite{duvsek2016sequence} adopts an end-to-end generation method that converts a meaning representation into a sentence. 
As attention mechanism   \cite{bahdanau2014neural} presents advantages in soft-searching the most relevant information among a sequence in neural machine translation task,  Nallapati et al. \cite{nallapati2016abstractive}
proposed a sequence-to-sequence attentional model to tackle text summarization task.
See at al. \cite{see2017get} proposes a hybrid pointer-generator network facilitating copying words from the source text via pointing \cite{vinyals2015pointer} while retaining the ability to produce new words via generator, and uses coverage to discourage repetition. 
Yu et al. \cite{yu2017seqgan} proposed SeqGAN framework that introduces GAN discriminator   \cite{goodfellow2014generative} to provide the reward signal and uses policy gradient technique \cite{sutton2000policy} to bypass the generator differentiation problem.
Lu et al. \cite{lu2018cot} proposed Cooperative Training (CoT) that coordinately trains a generative module and an auxiliary predictive module, to optimize the estimated Jensen-Shannon divergence. 

Besides the studies on design and training of language models, the researchers also proposed many indicators for evaluating the quality of samples generated by language models. 
These metrics can be classified into word-based metrics and grammar-based metrics. Word-based metrics move from simple $n$-gram overlap (including BLEU, TER   \cite{snover2006study}, ROUGE   \cite{lin2004rouge}, NIST   \cite{doddington2002automatic}, LEPOR   \cite{han2012lepor}, CIDER   \cite{vedantam2015cider} and METEOR   \cite{banerjee2005meteor}) to semantic similarity like Semantic Text Similarity   \cite{han2013umbc_ebiquity}. Grammar-based metrics include F-score, MaxMatch   \cite{dahlmeier2012better}, I-measure   \cite{felice2015towards}. 
Besides, instead of comparing sentences words by words, EmbSim   \cite{zhu2018texygen} compares the word embeddings. Some metrics are likelihood-based metrics that estimate the cross-entropy between the generated sentences and the true data, such as $\text{NLL}_{\text{oracle}}$   \cite{yu2017seqgan} that estimates average negative log-likelihood of generated sentences on oracle LSTM.

\section{Conclusion}
\label{Conclusion}
In this paper, we studied the problem of converting knowledge base RDF triples into natural languages. To handle this problem, we formulated it as a conditional natural language problem and utilized the discrete sequence generative models. We analyzed the limitations of existing methods on conditional sequence generative models and proposed a new method T2T which approximately optimizes an inverse Kullback-Leibler divergence between the real distribution and the learned one. We validated the proposed method on three benchmark datasets. The experiment results show that our method outperforms the baselines.

Our model is not limited in the task of translating knowledge bases RDF triples to natural languages; it can also be applied to other conditional generation tasks like machine translation and question answering systems, which we leave as future work.

\begin{acks}
The work is sponsored by Huawei Innovation Research Program. The corresponding author Weinan Zhang thanks the support of National Natural Science Foundation of China (61702327, 61772333, 61632017), Shanghai Sailing Program (17YF1428200). Any opinions, findings and conclusions or recommendations expressed in this material are those of the authors and do not necessarily
reflect those of the sponsor.

\end{acks}

\bibliographystyle{ACM-Reference-Format}
\bibliography{sample-base}


\end{document}